\documentclass[letterpaper, 10 pt, conference]{ieeeconf}  

\IEEEoverridecommandlockouts                              

\usepackage{amsmath} 
\usepackage{amssymb}  
\usepackage{balance}
\usepackage{graphicx}
\usepackage{times}
\usepackage{dsfont}
\usepackage{multicol}
\usepackage{siunitx}
\usepackage[bookmarks=true]{hyperref}
\usepackage{cleveref}
\usepackage{xspace}
\usepackage{xurl}
\usepackage{capt-of}
\usepackage[usenames,dvipsnames]{xcolor}
\usepackage{booktabs}  

\usepackage[
    style=ieee,
    natbib=true,
    citestyle=numeric-comp,
    doi=false,
    isbn=false,
    url=false]{biblatex} 

\definecolor{mydarkblue}{rgb}{0,0.08,0.45}
\definecolor{mydarkgreen}{RGB}{0, 139, 69}
\definecolor{mygreen2}{RGB}{0 205 0}
\definecolor{mybrown}{RGB}{139 69 19}

\definecolor{boxblue}{RGB}{79,173,234}
\definecolor{boxgreen}{RGB}{159,206,99}
\hypersetup{
	colorlinks=true,
	linkcolor=blue,
	urlcolor=magenta,
	citecolor=mygreen2,
}

\usepackage{amsfonts}
\usepackage{bm}
\usepackage{amsmath}
\usepackage{mathtools}
\usepackage{multirow}
\usepackage{booktabs}
\usepackage{caption}
\usepackage{wrapfig,lipsum,booktabs}
\usepackage{caption}
\usepackage{bbm}
\usepackage{multirow}
\usepackage{pifont}
\usepackage[linesnumbered,ruled,vlined]{algorithm2e}

\SetCommentSty{mycommfont}
\SetAlCapFnt{\small} 
\usepackage{etoc}
\SetAlgorithmName{Algo}{Algo}{}

\renewcommand{\paragraph}[1]{{\vspace{1mm}\noindent \bf #1}.}

\newcommand{\policyim}{{\pi_{\text{privileged}}}}

\newcommand{\reft}{{\bs{\hat{p}}_{t}}}
\newcommand{\reftkp}{{\bs{\hat{p}}_{t}}^\text{kp}}
\newcommand{\simtkp}{{\bs{{p}}_{t}}^\text{kp}}

\newcommand{\refr}{{\bs{\hat{\theta}}_{t}}}

\newcommand{\reflvkp}{{\bs{\hat{\dot{p}}}_{t}}^\text{kp}}

\newcommand{\refav}{{\bs{\hat{\omega}}_{t}}}
\newcommand{\reflv}{{\bs{\hat{v}}_{t}}}

\newcommand{\reftn}{{\bs{\hat{p}}_{t+1}}}
\newcommand{\refrn}{{\bs{\hat{\theta}}_{t+1}}}

\newcommand{\torque}{{\bs{{\tau}}_{t}}}
\newcommand{\dofpos}{{\bs{{q}}_{t}}}
\newcommand{\refdofpos}{{\bs{\hat{{q}}_{t}}}}
\newcommand{\rootvel}{{\bs{{v}}_{t}}}
\newcommand{\rootangvel}{{\bs{\omega}_{t}}}
\newcommand{\gravity}{{\bs{g}_{t}}}
\newcommand{\goalstateprivileged}{{\bs{s}^{\text{g-privileged}}_t}}
\newcommand{\selfstateprivileged}{{\bs{s}^{\text{p-privileged}}_t}}

\newcommand{\dofvel}{{\bs{\dot{q}}_{t}}}
\newcommand{\refdofvel}{{\bs{\hat{\dot{q}}}_{t}}}
\newcommand{\dofacc}{{\bs{\ddot{q}}_{t}}}

\newcommand{\simav}{{\bs{{\omega}}_{t}}}
\newcommand{\simlv}{{\bs{v}_{t}}}
\newcommand{\trans}{{\bs{p}}}

\newcommand{\simr}{{\bs{{\theta}}_{t}}}
\newcommand{\simt}{{\bs{{p}}_{t}}}
\newcommand{\goalstate}{{\bs{s}^{\text{g}}_t}}
\newcommand{\goalstateredu}{{\bs{s}^{\text{g-reduced}}_t}}

\newcommand{\selfstate}{{\bs{s}^{\text{p}}_t}}
\newcommand{\state}{{\bs{s}_t}}
\newcommand{\action}{{\bs{a}_t}}
\newcommand{\actionprev}{{\bs{a}_{t-1}}}
\newcommand{\motiondata}{\bs{\hat{Q}}}
\newcommand{\motiondataretarget}{\bs{\hat{Q}}^{\text{retarget}}}
\newcommand{\motiondataclean}{\bs{\hat{Q}}^{\text{clean}}}

\newcommand{\smplpose}{{\bs{{\theta}}}}
\newcommand{\smplshape}{{\bs{{\beta}}}}

\newcommand{\bs}[1]{\boldsymbol{#1}}
\newcommand{\xmark}{\ding{55}}%

\newcommand{\method}{\texttt{H2O}\xspace}
\newcommand{\methodwosim}{\method-w/o-sim2data\xspace}
\newcommand{\methodpo}{\method-reduced\xspace}

\title{\LARGE \bf
Learning Human-to-Humanoid Real-Time Whole-Body Teleoperation
}

\author{\authorblockN{Tairan He\textsuperscript{\dag}
\quad Zhengyi Luo\textsuperscript{\dag} \quad Wenli Xiao \quad Chong Zhang \quad Kris Kitani \quad Changliu Liu \quad Guanya Shi}
\authorblockA{
Carnegie Mellon University \\
\textsuperscript{\dag}Equal Contributions \quad
\href{https://human2humanoid.com}{\texttt{https://human2humanoid.com}}
}
}

\begin{document}

\makeatletter
\let\@oldmaketitle\@maketitle
    \renewcommand{\@maketitle}{\@oldmaketitle
    \centering
    \includegraphics[width=1.0\textwidth]{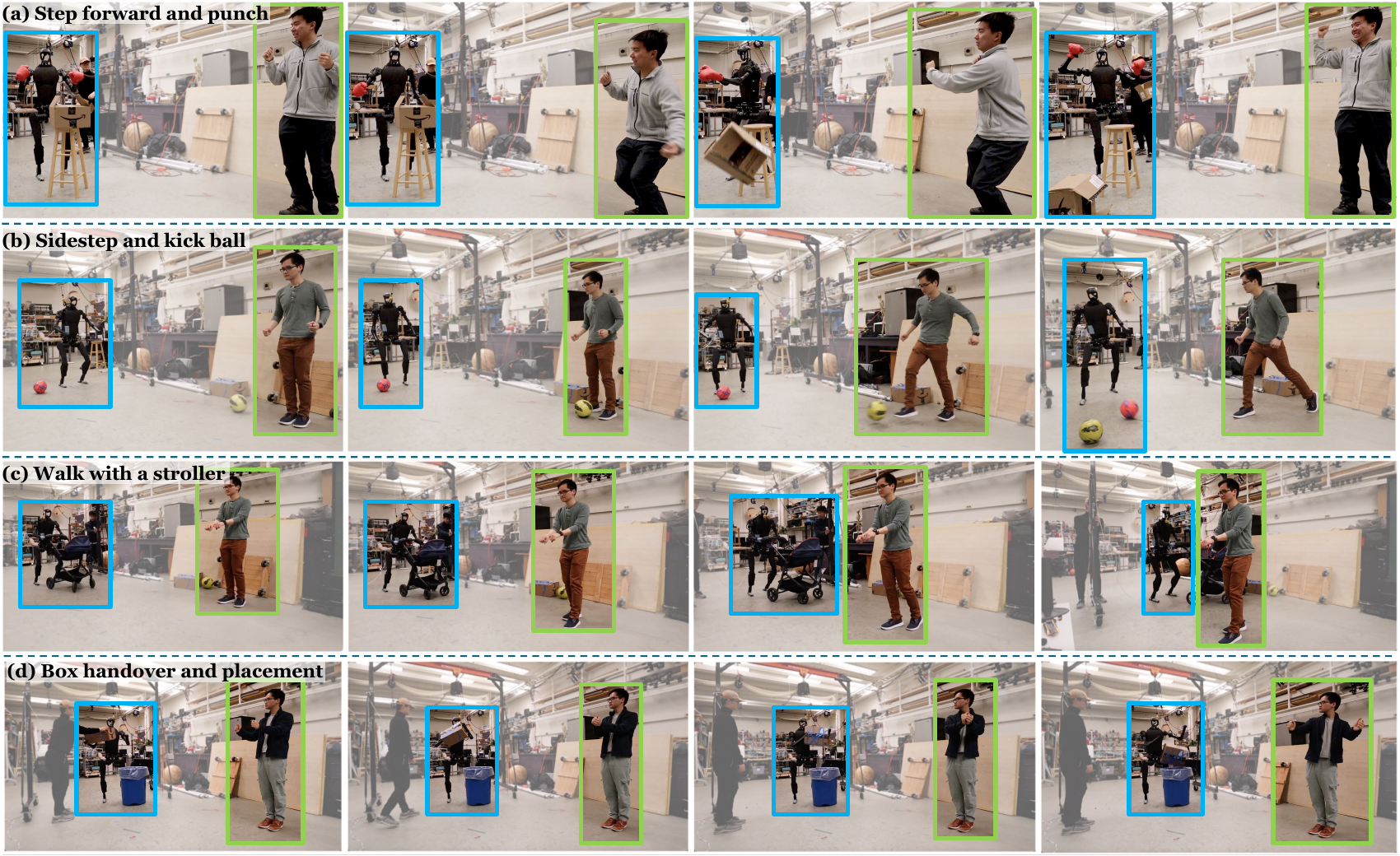}
    \vspace{-0.5cm}
    \captionof{figure}{
    The \textcolor{boxblue}{\textbf{humanoid robot}} is teleoperated in real-time using an RGB camera by the \textcolor{boxgreen}{\textbf{human teleoperator}}.
    (a) The humanoid mimics the human teleoperator, advancing one step while delivering a punch to displace a box, followed by a victory gesture.
    (b) The humanoid executes a precise sidestep to align with a ball and delivers a controlled kick using its right foot.
    (c) The humanoid demonstrates forward walking while pushing a stroller.
    (d) The operator teleoperates the humanoid to catch a box, rotate its waist, and drop the box into a waste bin. \textbf{Videos:} see the  \href{https://human2humanoid.com}{website}.}
    \vspace{-0.2cm} 
    \label{fig:firstpage}
    \setcounter{figure}{1}
    \vspace{-2mm}
  }
\makeatother

\maketitle

\pagestyle{empty}

\begin{abstract}
We present \textcolor{BrickRed}{H}uman \textcolor{BrickRed}{to} Human\textcolor{BrickRed}{o}id (\method), a reinforcement learning (RL) based framework that enables real-time whole-body teleoperation of a full-sized humanoid robot with only an RGB camera. 
To create a large-scale retargeted motion dataset of human movements for humanoid robots, we propose a scalable ``sim-to-data" process to filter and pick feasible motions using a privileged motion imitator.
Afterwards, we train a robust real-time humanoid motion imitator in simulation using these refined motions and transfer it to the real humanoid robot in a zero-shot manner. 
We successfully achieve teleoperation of dynamic whole-body motions in real-world scenarios, including walking, back jumping, kicking, turning, waving, pushing, boxing, \etc. 
To the best of our knowledge, this is the first demonstration to achieve learning-based real-time whole-body humanoid teleoperation.
\end{abstract}

\section{Introduction}
We aim to enable real-time teleoperation of a full-sized humanoid robot by a human teleoperator using an RGB camera. Humanoid robots, with their physical form closely mirroring that of humans, present an unparalleled opportunity for real-time teleoperation. This alignment of the embodiment allows for a seamless integration of human cognitive skills with versatile humanoid capabilities~\cite{darvish2023teleoperation}. 
Such synergy stimulated by human-to-humanoid teleoperation is crucial for complex tasks (\eg, household chores, medical assistance, high-risk rescue operations) that are yet too challenging for a fully autonomous robot, but possible for existing hardware teleoperated by humans~\cite{fu2024mobile, chi2024universal}.
In this paper, we transfer human motions to humanoid behaviors in a real-time fashion using an RGB camera. This system also has the potential to enable large-scale and high-quality data collection of human operations for robotics~\cite{zhao2023learning,fu2024mobile}, where human-teleoperated actions can be used for imitation learning. 

However, whole-body control of full-sized humanoids is a long-standing problem in robotics~\cite{kulic2016anthropomorphic}, and complexity increases when controlling the humanoid to replicate free-form human movements in real-time~\cite{darvish2023teleoperation}. Existing work on whole-body humanoid teleoperation has achieved remarkable results via model-based controllers~\cite{montecillo2010real, ishiguro2018high, ramos2018humanoid, ishiguro2020bilateral}, but they all use simplified models due to the high computational cost of modeling the full dynamics of the system~\cite{yamane2010controlling, zhang2023slomo}, which limits the scalability to dynamic motions. Furthermore, these works are highly dependent on contact measurement~\cite{di2016multi, otani2017adaptive}, leading to reliance on external setups such as the exoskeleton~\cite{ishiguro2020bilateral} and force sensors~\cite{ramos2018humanoid, ramos2019dynamic} for teleoperation.

Recent advances in reinforcement learning (RL) for humanoid control provide a promising alternative. First, in the graphics community, RL has been used to generate complex human movements~\cite{peng2018deepmimic, won2020scalable}, perform a variety of tasks \cite{peng2022ase}, and track real-time human motions captured by a webcam \cite{Luo_2023_ICCV} in simulation. However, due to unrealistic state-space design and partial disregard of the hardware limit (\eg torque / joint limit), it remains a question whether these methods can be applied to a full-sized humanoid. 
On the other hand, RL has achieved robust and agile biped locomotion in the real world~\cite{li2024reinforcement, radosavovic2023learning, siekmann2021blind}. 
To date, however, there has been no existing work on RL-based whole-body humanoid teleoperation.
The most closely related effort is a concurrent study~\cite{cheng2024expressive}, which focuses on learning to replicate upper-body motions and uses root velocity tracking for the lower body, from offline human motions rather than real-time teleoperation.

In this paper, we design a complete system for humanoid teleportation in real time. First, we identify one of the primary challenges in whole-body humanoid teleoperation as the lack of a dataset with feasible motions tailored to the humanoid, which is essential for training a controller that can track diverse motions. Although direct human-to-humanoid retargeting has been explored in previous locomotion-focused efforts~\cite{darvish2019whole,cisneros2024cybernetic, radosavovic2024humanoid}, retargeting a large-scale human motion dataset to the humanoid presents new challenges. That is, the significant dynamics discrepancy between humans and humanoids means that some human motions could be infeasible for the humanoid (\eg cartwheeling, steps wider than the leg lengths of the humanoid). In light of this, we introduce an automated ``sim-to-data" process to retarget and refine a large-scale human motion dataset \cite{mahmood2019amass} into motions that are feasible for real-world humanoid embodiment. Specifically, we first retarget the human motions to the humanoid via inverse kinematics, and train a humanoid controller with access to privileged state information~\cite{Luo_2023_ICCV} to imitate the unfiltered motions in simulation. Afterwards, we remove the motion sequences that the privileged imitator fails to track. By doing so, we create a large-scale humanoid-compatible motion dataset.

After obtaining a dataset of feasible motions, we develop a scalable training process for the real-world motion imitator that incorporates extensive domain randomization to bridge the sim-to-real gap. To facilitate real-time teleoperation, we design a state space that prioritizes the inputs available in the real world using an RGB camera, such as the keypoint positions. During inference, we use an off-the-shelf human pose estimator~\cite{li2021hybrik} to provide global human body positions for the humanoid to track.

In summary, we demonstrate the feasibility of an RL-based real-time \textcolor{BrickRed}{H}uman-\textcolor{BrickRed}{to}-Human\textcolor{BrickRed}{o}id (\method) teleoperation system. Our contributions include:
\begin{enumerate}
    \item A scalable retargeting and ``sim-to-data" process to obtain a large-scale motion dataset feasible for the real-world humanoid robot;
    \item Sim-to-real transfer of the RL-based whole-body tracking controller that scales to a large number of motions;
    \item A real-time teleoperation system with an RGB camera and 3D human pose estimation, demonstrating fulfillment of various whole-body motions including walking, pick-and-place, stroller pushing, boxing, hand-waving, ball kicking, etc.
\end{enumerate}

\section{Related Works}

\subsection{Physics-Based Animation of Human Motions}

Physics-based simulation has been used to generate realistic and natural motions for avatars~\cite{peng2017deeploco, peng2018deepmimic, wang2020unicon, won2020scalable, fussell2021supertrack, peng2021amp, peng2022ase, luo2024universal, Luo_2023_ICCV, winkler2022questsim}. With motion capture as the main source of human motion data~\cite{mahmood2019amass}, RL is often used to learn avatar controllers that can mimic these motions, offering distinctive styles~\cite{peng2018deepmimic, peng2021amp}, scalability~\cite{won2020scalable, luo2021dynamics, Luo_2023_ICCV}, and reusability~\cite{peng2022ase, luo2024universal}.

However, realistic animation in physics-based simulators does not guarantee real-world applicability, especially for humanoids. Simulated humanoid avatars typically have high degrees of freedom and large joint torques~\cite{humanoid_urdf}, and sometimes need non-physical assistive external forces~\cite{yuan2020residual}. In this work, we demonstrate that, with carefully designed sim-to-real training, approaches in the humanoid animation community can be applied to a real-world humanoid robot.

\subsection{Transferring Human Motions to Real-World Humanoids}
Before the emergence of RL-based humanoid controllers, traditional methods typically employ model-based optimization to track retargeted motions while maintaining stability~\cite{darvish2023teleoperation}. To this end, these methods minimize tracking errors under the constraints of stability and contacts, requiring predefined contact states~\cite{montecillo2010real, di2016multi, yamane2010controlling, otani2017adaptive,penco2018robust, koenemann2014real, ramos2015dancing, penco2024mixed} or estimated contacts from sensors~\cite{ayusawa2017motion, ramos2019dynamic, hu2014online, ishiguro2018high, ishiguro2020bilateral}, hindering large-scale deployment outside the laboratory. \citet{zhang2023slomo} use contact-implicit model predictive control (MPC) to track motions extracted from videos, but trajectories must first be optimized offline to ensure dynamic feasibility. Furthermore, the model used in MPC needs to be simplified due to computational burden~\cite{montecillo2010real, ramos2019dynamic, zhang2023slomo}, which limits the capability of trackable motions.

RL-based controllers may provide an alternative that does not require explicit contact information. Some works~\cite{bohez2022imitate, tang2023humanmimic} use imitation learning to transfer human-style motions to the controller, but do not accurately track human motions. \citet{cheng2024expressive} train whole-body humanoid controllers that can replicate upper body movements from offline human motions, but the lower body relies on root velocity tracking and does not track precise lower body movements. In comparison, our work achieves real-time whole-body tracking of human motions.

\subsection{Teleoperation of Humanoids}

Teleoperation of humanoids can be categorized into three types: 1) task-space teleoperation~\cite{seo2023deep, dafarra2024icub3}, 2) upper-body-retargeted teleoperation~\cite{chagas2021humanoid, elobaid2020telexistence}, and 3) whole-body teleoperation~\cite{montecillo2010real, otani2017adaptive, hu2014online, tachi2020telesar, ishiguro2018high}. 
For the first and second types, the shared morphology between humans and humanoids is not fully utilized, and whole-body control must be solved in a task-specified way. This also raises the concern that, if tracking lower body movement is not necessary, the robot could opt for designs with better stability, such as a quadruped~\cite{bellicoso2019alma} or wheeled configuration~\cite{lenz2023nimbro}.

Our work belongs to the third type and is the first to achieve learning-based whole-body teleoperation. 
Moreover, our approach does not require capture markers or force sensors on the human teleoperator, as we directly employ an RGB camera to capture human motions for tracking, potentially paving the way for collecting large-scale humanoid data for training autonomous agents.

\section{Preliminaries}
The whole-body real-time humanoid teleoperation we are tackling is formulated as a goal-conditioned RL problem, which tracks versatile human motion with a single RL control policy. In \Cref{sec:gcrl}, we set up the preliminary for our control framework. \Cref{sec:human-model} describes the human model and dataset we use in the RL policy training. As a notation convention, we use $\widetilde{\cdot}$ to represent kinematic quantities (without physics simulation) from pose estimator/keypoint detectors, $\widehat{\cdot}$ to denote ground truth quantities from Motion Capture (MoCap), and normal symbols without accents for values from the physics simulation. 


\subsection{Goal-conditioned RL for Humanoid Control}
\label{sec:gcrl}
We formulate our problem as goal-conditioned RL where $\pi$ is trained to track real-time human motions. We formulate the learning task as a Markov Decision Process (MDP) defined by the tuple $\mathcal{M}=\langle\mathcal{S}, \mathcal{A}, \mathcal{T}, \mathcal{R}, \gamma\rangle$ of state $\state \in \mathcal{S}$, action $\action \in \mathcal{A}$, transition dynamics $\mathcal{T}$, reward function $\mathcal{R}$, and discount factor $\gamma$. The state $\state$ contains the proprioception $\selfstate$ and the goal state $\goalstate$. The goal state $\goalstate$ is a unified representation of the whole-body motion of the human teleoperator, which we will discuss in detail in \Cref{sec:obs-space}. Based on proprioception $\selfstate$ and goal state $\goalstate$, we define the reward $r_t=\mathcal { R }\left(\selfstate, \goalstate \right)$ for the policy training. The action $\action \in \mathbb{R}^{19}$ specifies the joint target positions that the PD controller will use to actuate the degrees of freedom.  We apply the proximal policy gradient (PPO)~\cite{SchulmanWDRK17} to maximize the cumulative discounted reward $\mathbb{E}\left[\sum_{t=1}^T \gamma^{t-1} r_t\right]$.  We formulate the teleoperation task as the motion imitation/tracking/mimicking task, where we train the humanoid to track the reference motion at every frame.

\subsection{Parametric Human Model and Human Motion Dataset}
\label{sec:human-model}
Popular in the vision and graphics community, parametric human models such as SMPL \cite{SMPL:2015} are easy to work with representations of human shapes and motions. SMPL represents the human body as body shape parameters $\smplshape \in \mathcal{R}^{10}$, pose parameters $\smplpose \in \mathcal{R}^{24 \times 3}$, and root translation $\trans \in \mathcal{R}^{24 \times 3}$. Given $\smplshape$,  $\smplpose$ and $\trans$, $\mathcal{S}$ denotes the SMPL function, where $\mathcal{S}(\smplshape, \smplpose, \trans): \smplshape, \smplpose, \trans \rightarrow \mathcal{R}^{6890\times3}$ maps the parameters to the position of the vertices of a triangular human mesh of 6890 vertices. The AMASS \cite{mahmood2019amass} dataset contains 40 hours of motion capture expressed in the SMPL parameters. 

\begin{figure}[tbp]
    \centering
    \includegraphics[width=0.5\textwidth]{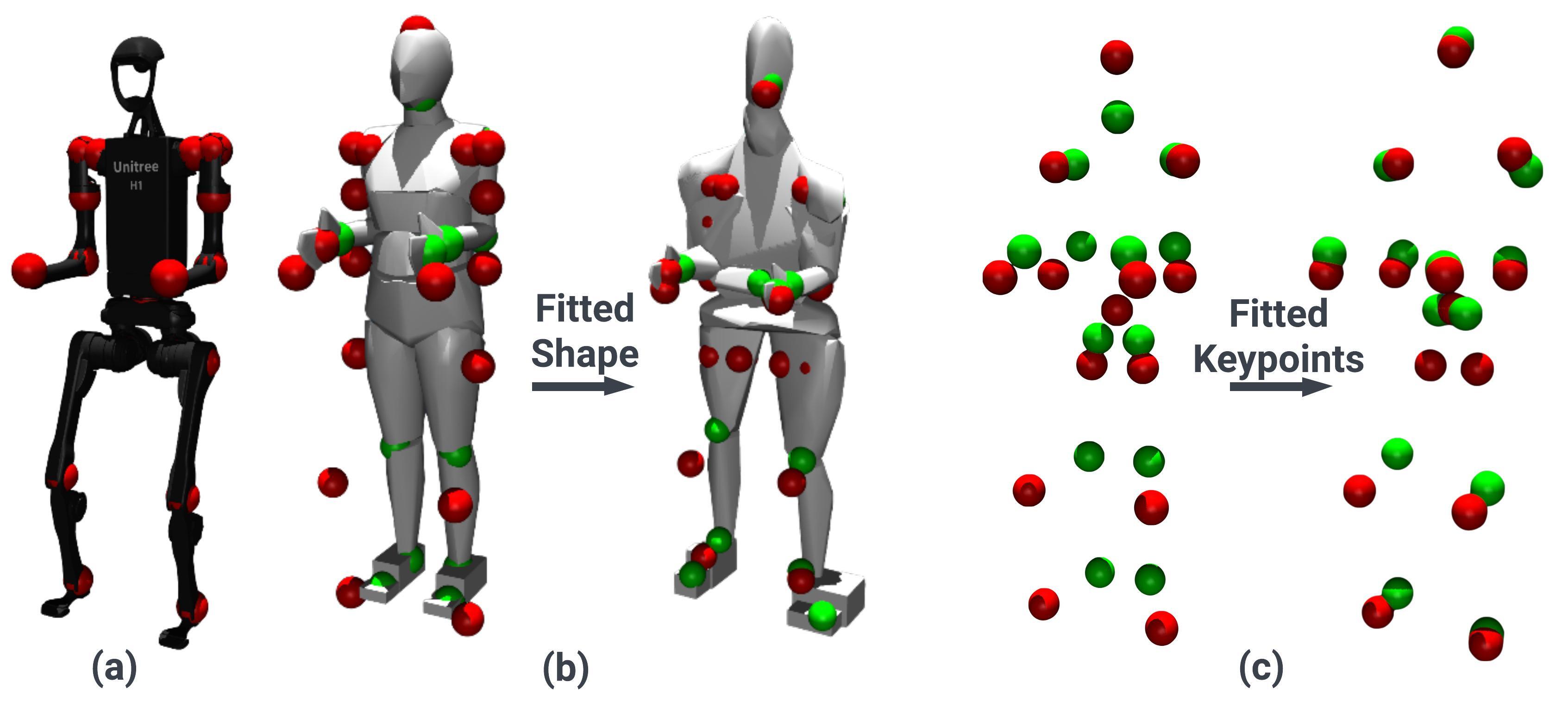}
    \caption{Fitting the SMPL body to the H1 humanoid. (a) Visualization of the humanoid keypoints (red dots) (b) Humanoid keypoints vs SMPL keypoints (green dots and mesh) before and after fitted SMPL shape $\smplshape'$.  (c) Corresponding 12 joint position before and after fitting. }
    \label{fig:H2O_fitting}
    \vspace{-2mm}
\end{figure}

\begin{figure}[tbp]
    \centering
    \includegraphics[width=0.4\textwidth]{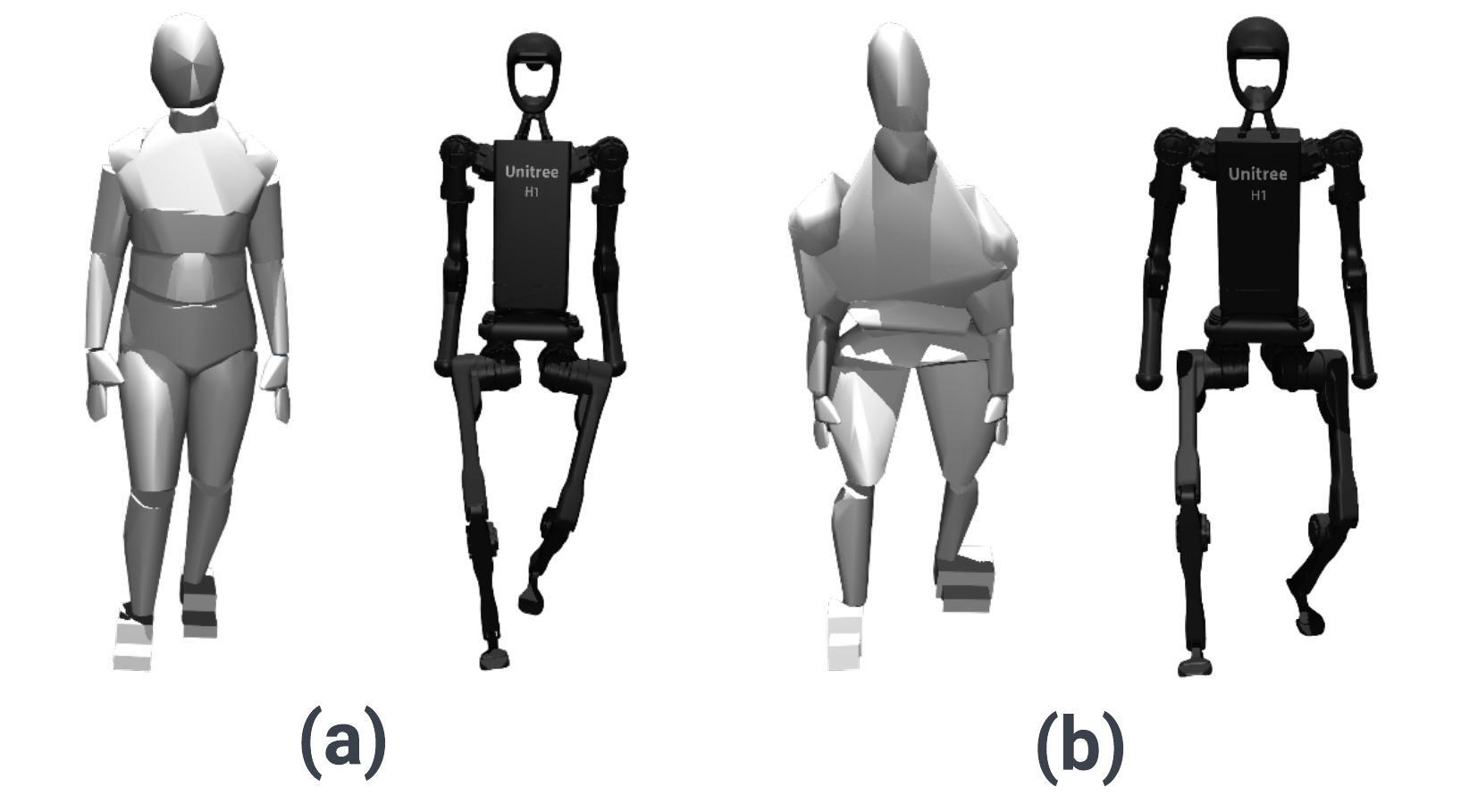}
    \caption{Effect of using a fitted SMPL shape $\smplshape'$ instead of mean body shape on position-based retargeting. (a) Retargting without using $\smplshape'$, which results in unstable ``in-toed" humanoid motion. (b) Retargeting using $\smplshape'$, which result in balanced humanoid motion. }
    \label{fig:H2O_shape}
    \vspace{-2mm}
\end{figure}

\section{Retargeting Human Motions for Humanoid}
\label{SEC:retargeting}
To enable humanoid motion imitation for unscripted human motion, we require a large amount of whole-body motion to train a robust motion imitation policy. Since humans and humanoids also have a nontrivial difference in body structure, shape, and dynamics, naively retargeted motion from a human motion dataset can result in a large number of motions impossible for our humanoid to perform. These imfeasible motion sequences can hinder imitation training as observed in prior work \cite{luo2024universal}. To resolve these issues, we design a ``sim-to-data'' approach to complement traditional retargeting to convert a large-scale human motion dataset to feasible motions for humanoids.

\begin{figure*}[t]
\centering\includegraphics[width=\textwidth]{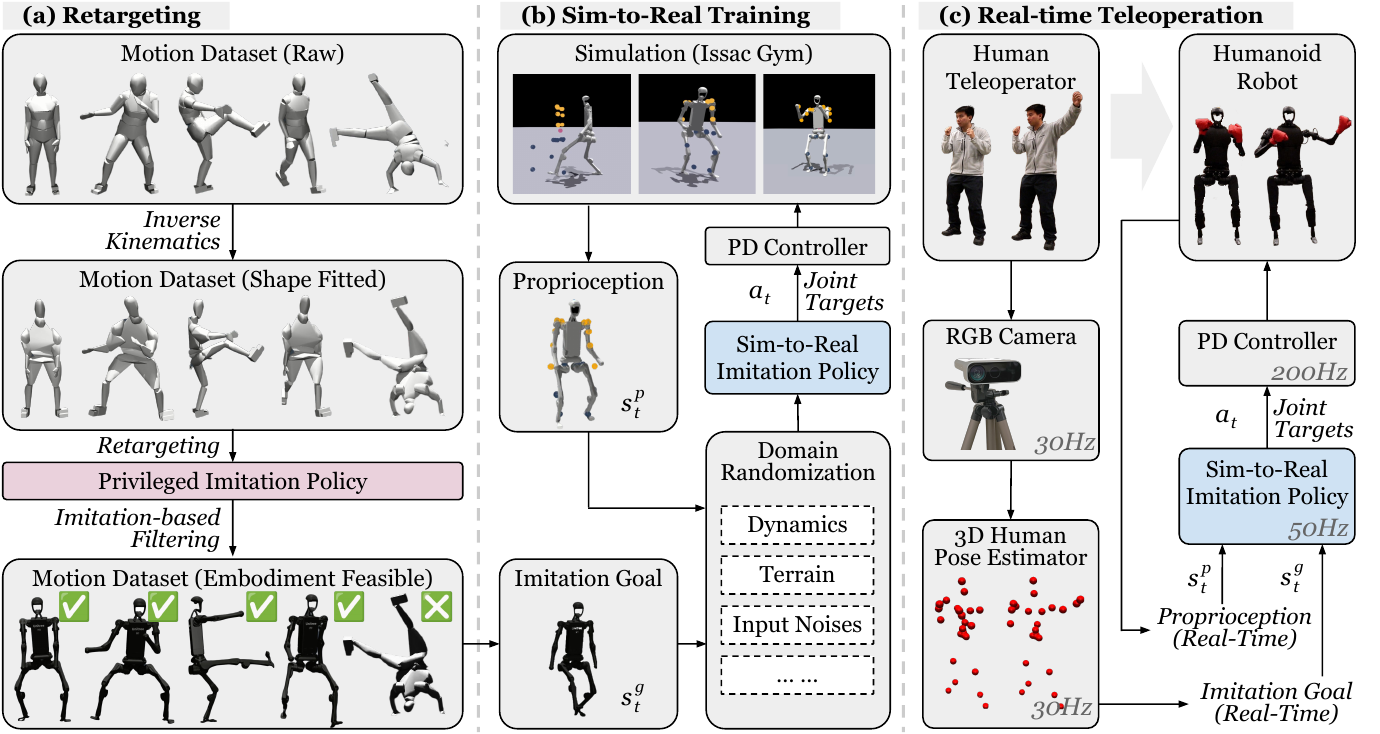}
    \caption{Overview of \method: (a) \textbf{Retargeting} (\Cref{SEC:retargeting}): \method first aligns the SMPL body model to a humanoid's structure by optimizing shape parameters. Then \method retargets and removes the infeasible motions using a trained privileged imitation policy, producing a clean motion dataset. (b) \textbf{Sim-to-Real Training}: (\Cref{SEC:PolicyTraining}) An imitation policy is trained to track motion goals sampled from a cleaned dataset. (c) \textbf{Real-time Teleoperation Deployment} (\Cref{sec:real_experiments}): The real-time teleoperation deployment captures human motion through an RGB camera and a pose estimator, which is then mimicked by a humanoid robot using the trained sim-to-real imitation policy.}
    \label{fig:H2O_overview}
    \vspace{-2mm}
\end{figure*}

\subsection{Motion Retargeting}
As there is a non-trivial difference between the SMPL kinematic structure and the humanoid kinematic tree, we perform a two-step process for the initial retargeting. First, since the SMPL body model can represent different body proportions, we first find a body shape $\smplshape'$ closest to the humanoid structure. We choose 12 joints that have a correspondence between humans and humanoids, as shown in Fig.\ref{fig:H2O_fitting} and perform gradient descents on the shape parameter $\bs{s}$ to minimize the joint distances using a common rest pose. After finding the optimal $\smplshape'$, given a sequence of motions expressed in SMPL parameters, we use the original translation $\trans$ and pose $\smplpose$, but the fitted shape $\smplshape'$ to obtain the set of body keypoint positions. Then we retarget motion from human to humanoid by minimizing the 12 joint position differences using Adam optimizer \cite{kingma2014adam}. Notice that our retargeting process try to match the end effectors of the human to the humanoid (\eg ankles, elbows, wrists) to preserve the overall motion pattern. Another approach is direct copying the local joint angles from human to humanoid, but that approach can lead to large differences in end-effector positions due to the large difference in kinematic trees. During this process, we also add some heuristic-based filtering to remove unsafe sequences, such as sitting on the ground. The motivation to find $\smplshape'$ before retargeting is that in the rest pose, our humanoid has a large gap between its feet. If naively trying to match the foot movement between the human and the humanoid, the humanoid motion can have an in-toed artifact. Using $\smplshape'$, we can find a human body structure has a large gap between its rest pose (as shown in Fig.\ref{fig:H2O_fitting}). Using $\smplshape'$ during fitting can effectively create motion that is more feasible for the humanoid, as shown in Fig.\ref{fig:H2O_shape}. From the AMASS dataset $\motiondata$ that contains 13k motion sequences, this process computes 10k retargted motion sequences $\motiondataretarget$.

\subsection{Simulation-based Data Cleaning}
As shown in \Cref{fig:H2O_overview}$, \motiondataretarget$ contain a large number of implausible motions for the humanoid due to the significant gap between the capabilities of a human and a motor-actuated humanoid. Manually finding these data sequences from a large-scale dataset can be a rather cumbersome process. Thus, we propose a ``sim-to-data" procedure, where we train a motion imitator $\policyim$ (similar to PHC \cite{Luo_2023_ICCV}) with access to privileged information and no domain randomization to imitate all uncleaned data $\motiondataretarget$. Without domain randomization, $\policyim$ can perform well in motion imitation, but is not suitable for transfer to the real humanoid. However, $\policyim$ represents the upper bound of motion imitation performance, and sequences which $\policyim$ fails to imitate represent implausible ones. Specifically, we train $\policyim$ following the same state space, control parameters, and hard-negative mining procedure proposed in PULSE \cite{luo2024universal}, and train a single imitation policy to imitate the entire retargeted dataset. After training, $\sim$8.5k out of 10k motion sequences from AMASS turn out to be plausible for the H1 humanoid, and we denote the obtained clean dataset as $\motiondataclean$.

\paragraph{Privileged Motion Imitation Policy} To train $\policyim$, we follow PULSE \cite{luo2024universal} and train a motion imitator with access to the full rigid body state of the humanoid. Specifically,  for the privileged policy $\policyim$, its proprioception is defined as $\selfstateprivileged \triangleq [\simt, \simr, \simlv, \simav]$, which contains the global 3D rigid body position $\simt$, orientation $\simr$, linear velocity $\simlv$, and angular velocity $\simav$ of all rigid bodies in the humanoid. The goal state is defined as $\goalstateprivileged \triangleq   [\refrn \ominus \simr, \reftn -  \simt, \bs{\hat{v}}_{t+1} -  \bs{v}_t, \bs{\hat{\omega}}_t -  \bs{\omega}_t, \refrn, \reftn]$, which contains the one-frame difference between the reference and current simulation result for all rigid bodies on the humanoid. It also contains the next frame's reference rigid body orientation and position. All values are normalized to the humanoid's coordinate system. Notice that all values are global, and values such as global rigidbody linear velocity $\simlv$ and angular velocity $\simav$ are hard to obtain accurately in the real world.

\label{SEC:Retargeting}

\section{Whole-Body Teleoperation Policy Training}
\label{SEC:PolicyTraining}
\subsection{State Space}
\label{sec:obs-space}
To achieve real-time teleoperation of humanoid robots, the state space of RL policy must contain only quantities available in the real world. This differs from the simulation-only approaches, where all the physics information (\eg, foot contact force) is available. For example, in the real world, we have no access to each joint's precise global angular velocity due to the lack of IMUs, but the privileged policy $\policyim$ requires them. 

In our state space design, proprioception is defined by $\selfstate \triangleq \left[\dofpos, \dofvel, \rootvel, \rootangvel, \gravity, \actionprev\right]$, with joint position $\dofpos\in\mathbb{R}^{19}$ (DoF position), joint velocity $\dofvel\in\mathbb{R}^{19}$ (DoF velocity), root linear velocity $\rootvel\in\mathbb{R}^3$, root angular velocity $\rootangvel\in\mathbb{R}^3$, root projected gravity $\gravity\in\mathbb{R}^3$, and last action $\actionprev\in\mathbb{R}^{19}$. The goal state is $\goalstate \triangleq  [\reftkp, \reftkp - \simtkp, \reflvkp]$. $\reftkp \in \mathbb{R}^{8\times3}$ is the position of eight selected reference body positions (shoulders, elbows, hands, ankles); $\reftkp - \simtkp$ is the position difference between the reference joints and humanoid's own joints; $\reflvkp$ is the linear velocity of the reference joints. All values are normalized to the humanoid's own coordinate system. 
As a comparison, we also consider a reduced goal state $\goalstateredu \triangleq  (\reftkp)$, where only the reference position $\reftkp$ but not the position difference. The action space of the agile policy consists of 19-dim joint targets. A PD controller tracks these joint targets by converting them to joint torques: $\tau=K_p(\action-\dofpos)-K_d \dofvel$.

\subsection{Reward Design}
\label{sec:reward}
We formulate the reward function $r_t$ with the summation of three terms: 1) penalty; 2) regularization; and 3) task rewards, which are summarized in detail in~\Cref{tab:reward}. 
Note that while we only have eight selected body positions $\reftkp$ in our state space, we provide six \textit{full-body} reward terms for all joints (DoF position, DoF velocity, body position, body rotation, body velocity, body angular velocity) for the imitation task. These expressive rewards give more dense reward signals for efficient RL training.

\begin{table}[tbp]
\centering
\caption{Reward components and weights: penalty rewards for preventing undesired behaviors for sim-to-real transfer, regularization to refine motion, and task reward to achieve successful whole-body tracking in real-time.}

\begin{tabular}{  c  c  c  }
\hline
Term                  & Expression & Weight    \\ \hline
               &     Penalty       &           \\ \hline
Torque limits        &     $ \mathds{1}({\torque \notin [\bs{\tau}_{\min}, \bs{\tau}_{\max} ]})  $          & $-2e^{-1}$   \\
DoF position limits      &   $ \mathds{1}({\dofpos \notin [\bs{q}_{\min}, \bs{q}_{\max} ]})  $      & $-1e^2$    \\
Termination           &     $\mathds{1}_\text{termination}$       & $-2e^2$    \\ \hline
        &    Regularization       &           \\ \hline
DoF acceleration               &   $\lVert \dofacc \rVert_2^2$              & $-8.4e^{-6}$ \\
DoF velocity               &   $\lVert \dofvel \rVert_2^2$         & $-3e^{-3}$   \\
Action rate     &    $ \lVert \action - \actionprev \rVert_2^2  $     & $-9e^{-1}$   \\
Torque                & $\lVert\torque\rVert$     & $-9e^{-5}$   \\
Feet air time       &      $T_\text{air}-0.25$~\cite{rudin2022learning}    & $8e^2$     \\ 
Feet contact force   &     $\lVert F_\text{feet}\rVert_2^2$       & $-1e^{-1}$   \\
Stumble               &     $\mathds{1}(F_\text{feet}^{xy} > 5\times F_\text{feet}^z)$       & $-1e^3$    \\
Slippage              &    $\lVert \simlv^\text{feet} \rVert_2^2 \times \mathds{1} (F_\text{feet} \geq 1)$        & $-3e^1$    \\ \hline
                  &    Task Reward        &           \\ \hline
DoF position        &   $\exp(-0.25\lVert \refdofpos -\dofpos\rVert_2)$         & $2.4e^1$   \\
DoF velocity        &   $\exp(-0.25 \lVert \refdofvel- \dofvel\rVert_2^2)$          & $2.4e^1$   \\
Body position         &   $\exp (-0.5\lVert \simt-\reft\rVert_2^2) $      & $4e^1$     \\
Body rotation         &   $\exp(-0.1\lVert \simr \ominus \refr\rVert)$         & $1.6e^1$   \\
Body velocity         &    $\exp (-10.0\lVert \simlv-\reflv\rVert_2) $        & $6e^1$     \\
Body angular velocity &   $\exp (-0.01\lVert \simav-\refav\rVert_2) $         & $6e^1$     \\ \hline
\end{tabular}%
\label{tab:reward}
\end{table}

\subsection{Domain Randomization}
\label{sec:domainrandomization}

\begin{table}[tbp]
\centering
\caption{The range of dynamics randomization. Describing simulated dynamics randomization, external perturbation, and randomized terrain, which are important for sim-to-real transfer and boost robustness and generalizability.}
\label{table:dr}
\begin{tabular}{ c c }
\hline
Term           & Value                              \\ \hline
\multicolumn{2}{c}{\textbf{Dynamics Randomization}}  \\ \hline

Friction & $\mathcal{U}(0.2, 1.1)$            \\
Base CoM offset    & $\mathcal{U}(-0.1, 0.1) \text{m}$           \\
Link mass    & $\mathcal{U}(0.7, 1.3) \times \text{default} \ \text{kg}$            \\
P Gain             & $\mathcal{U}(0.75, 1.25) \times \text{default}$          \\
D Gain             & $\mathcal{U}(0.75, 1.25)  \times \text{default} $         \\
Torque RFI~\cite{campanaro2023learning}         & $0.1 \times \text{torque limit}\ \text{N}\cdot\text{m}$  \\
Control delay      & $\mathcal{U}(20, 60)\text{ms}$           \\ \hline
\multicolumn{2}{c}{\textbf{External Perturbation}}  \\ \hline
Push robot         & $\text{interval}=5s$, $v_{xy}=0.5 \text{m/s}$                   \\ \hline
\multicolumn{2}{c}{\textbf{Randomized Terrain}}  \\ \hline
Terrain type         & flat, rough, low obstacles~\cite{he2024agile}                  \\ \hline
\end{tabular}
\vspace{-3mm}
\label{tab:DR}
\end{table}

Domain randomization has been shown to be the key source of robustness and generalization to achieve successful sim-to-real transfers~\cite{peng2018sim,li2024reinforcement}. All the domain randomization we use in \method are listed in \Cref{tab:DR}, including ground friction coefficient, link mass, Centor-of-Mass (CoM) position of the torso link, PD gains of the PD controller, torque noise on the actually applied torques on each joint, control delay, terrain types. 
The link mass and PD gains are independently randomized for each link and joint, and the rest are episodic randomized.
These domain randomization together can effectively facilitate the sim-to-real transfer for the real-world dynamics and hardware gaps.

\subsection{Early Termination Conditions}
We introduce three early termination conditions to make the RL training process more sample-efficient: 1) low height: the base height is lower than 0.3m; 2) orientation: the projected gravity on x or y axis exceeds 0.7; 3) teleoperation tolerance: the average link distance between the robot and reference motions is further than 0.5m.

\section{Experimental Results}
\label{SEC:Experiment}

\begin{table*}[t]
\caption{Quantitative motion imtiation results the uncleaned retargeted AMASS dataset $\motiondataretarget$.}
\label{tab:amass_imitation}
\centering
\resizebox{\linewidth}{!}{%
\begin{tabular}{lccrrrrrrrrr}
\toprule
\multicolumn{3}{c}{} & \multicolumn{5}{c}{All sequences} & \multicolumn{4}{c}{Successful sequences}
\\ 
\cmidrule(r){1-3} \cmidrule(r){4-8} \cmidrule(r){9-12}
Method  & State Dimension & Sim2Real & $\text{Succ} \uparrow$ & $E_\text{g-mpjpe}  \downarrow$ &  $E_\text{mpjpe} \downarrow $ &  $\text{E}_{\text{acc}} \downarrow$  & $\text{E}_{\text{vel}} \downarrow$ & $E_\text{g-mpjpe}  \downarrow$ &  $E_\text{mpjpe} \downarrow $ &  $\text{E}_{\text{acc}} \downarrow$  & $\text{E}_{\text{vel}} \downarrow$   \\ 
\cmidrule(r){1-1} \cmidrule(r){2-3} \cmidrule(r){4-8} \cmidrule(r){9-12}
Privileged policy & $\mathcal{S}\subset\mathcal{R}^{778}$ &  \xmark \  & {85.5\%} &  {50.0} & {43.6} & {6.9} & {7.8} &  {46.0} & {40.9} & {5.2} & {6.2} \\
\cmidrule(r){1-1} \cmidrule(r){2-3} \cmidrule(r){4-8} \cmidrule(r){9-12}
\methodpo & $\mathcal{S}\subset\mathcal{R}^{90}$ & \checkmark  &{53.2\%} & {200.2} & {115.8} & {11.2} & {13.8} & {182.5} & {111.0} & {3.0} & {8.1}  \\
\methodwosim & $\mathcal{S}\subset\mathcal{R}^{138}$ & \checkmark & {67.9\%} & {176.6} & {95.0} & {10.2} & {12.2}  & {163.1} & {93.8} & {3.0} & {7.5} \\
\method & $\mathcal{S}\subset\mathcal{R}^{138}$ & \checkmark & \textbf{72.5\%} & \textbf{166.7} & \textbf{ 91.7}  & \textbf{8.9} & \textbf{11.0}  & \textbf{151.0} & \textbf{88.8}  & \textbf{2.9} & \textbf{7.0}\\

\bottomrule 
\end{tabular}}
\end{table*}

\subsection{Simulation Experiments}
\label{sec:simulaiton_experiment}

\paragraph{Baselines} To reveal the effect of different retargeting, state space designs, and sim-to-real training techniques on whole-body teleoperation performance, we consider four baselines:
\begin{enumerate}
    \item Privileged policy $\policyim$: The privileged policy (trained without any sim-to-real regularizations or domain randomizations) is used to filter the dataset to find infeasible motion. It has no sim-to-real capability and has a much higher input dimension. 
    \item \methodwosim: \method without the ``sim-to-data'' retargeting, trained on the $\motiondataretarget$;
    \item \methodpo: \method with a state space of goal state consisting only of selected body positions $\goalstateredu$.
    \item \method: Our full \method system, with all the retargeting process introduced in \Cref{SEC:retargeting} and the state space design introduced in \Cref{sec:obs-space}, trained on $\motiondataclean$;
\end{enumerate}

\paragraph{Metrics} We evaluate these baselines in simulation on the uncleaned retargeted AMASS dataset (10k sequences $\motiondataretarget$). The metrics are as follows:
\begin{enumerate}
    \item Success rate: the success rate (Succ) as in PHC~\cite{Luo_2023_ICCV}, deeming imitation unsuccessful when, at any point during imitation, the average difference in body distance is on average further than 0.5m. Succ measures whether the humanoid can track the reference motion without losing balance or significantly lag behind. 
    \item  $E_{\text{mpjpe}}$ and $E_{g-\text{mpjpe}}$: \text{the global MPJPE } $E_{g-\text{mpjpe}}$ and the root-relative mean per-joint position error (MPJPE)  $E_{\text{mpjpe}}$ (in mm), measuring our imitator’s ability to imitate the reference motion both globally and locally (root-relative). 
    \item $E_\text{acc}$ and $E_\text{vel}$: To show physical realism, we also compare acceleration  $E_{\text{acc}}$ $\text{(mm/frame}^2$ and velocity  $E_{\text{vel}}$ (mm/frame) difference. 
\end{enumerate}

\paragraph{Results} The experimental results are summarized in \Cref{tab:amass_imitation}, where \method significantly outperforms \methodwosim and \methodpo by a large margin, demonstrating the importance of the ``sim-to-data'' process and the state-space design of motion goals for RL. Note that the privileged policy and \methodwosim are trained on the entire retargeted AMASS dataset $\motiondataretarget$ while \method and \methodpo are trained on the filtered dataset $\motiondataclean$. 
The success rate gap between \method and the privileged policy comes from two factors: 1) \method uses a much more practical and less informative observation space compared to the privileged policy; 2) \method is trained with all sim-to-real regularizations and domain randomization. These two factors will both lead to degradation in simulation performance. This shows that while the RL-based avatar control frameworks have achieved impressive results in simulation, transferring them to the real world requires more robustness and stability. With the carefully chosen dataset and the state space, we could make \method achieve a higher success rate compared to \methodwosim and \methodpo. 
By comparing \method with \methodwosim, we can see that our ``sim-to-data'' process is effective in obtaining higher success rate, even when the RL policy is trained on \textit{less} data. Intuitively, an implausible motion may cause the policy to waste resources trying to achieve them, and filtering them out can lead to better overall performance, as also observed in PULSE \cite{luo2024universal}. Comparing \method with \methodpo, the only difference is the design of the state space of the goal, which indicates that including more informative physical information about motions helps RL to generalize to large-scale motion imitation.

\paragraph{Ablation on Motion Dataset Size}
To show how motion tracking performance scales with the size of the motion dataset, we test \method with different size of $\motiondataclean$ by randomly selecting $1\%$, $10\%$ of $\motiondataclean$. The results are summarized in \Cref{tab:scale_dataset_size}, where policies trained larger motion datasets continue to improve the tracking performance. Notice that a policy trained on only 0.1\% of the data can achieve a surprisingly high success rate, most likely due to the ample domain randomization applied to the humanoid, such as push robot significantly widens the state the humanoid has encountered, improving its generalization capability.

\begin{table}[t]
\caption{Quantitative results of \method on different sizes of motion dataset for training, evaluated on the uncleaned retargeted AMASS dataset $\motiondataretarget$.}
\label{tab:scale_dataset_size}
\centering
\resizebox{\linewidth}{!}{%
\begin{tabular}{lrrrrr}
\toprule
\multicolumn{1}{c}{} & \multicolumn{5}{c}{All sequences} 
\\ 
\cmidrule(r){1-1} \cmidrule(r){2-6}
Dataset Size & $\text{Succ} \uparrow$ & $E_\text{g-mpjpe}  \downarrow$ &  $E_\text{mpjpe} \downarrow $ &  $\text{E}_{\text{acc}} \downarrow$  & $\text{E}_{\text{vel}}$   \\ 
\cmidrule(r){1-1} \cmidrule(r){2-6}
$0.1\% \ \text{of} \ \motiondataclean$  & {52.0\%} & {198.0} & {107.9}  & {12.4} & {13.7}\\
$1\% \ \text{of} \  \motiondataclean$ & {58.8\%} & {183.8} & {96.4}  & {10.7} & {12.0}\\
$10\% \ \text{of} \ \motiondataclean$  &{61.3\%} & {174.3} & {92.3} & {10.8} & {12.1} \\
$100\% \ \text{of} \ \motiondataclean$  & \textbf{72.5\%} & \textbf{166.7} & \textbf{ 91.7}  & \textbf{8.9} & \textbf{11.0}\\

\bottomrule 
\end{tabular}}
\vspace{-2mm}
\end{table}

\subsection{Real-world Demonstrations}
\label{sec:real_experiments}
\paragraph{Deployment Details} For real-world deployment tests, we use a standard 1080P webcam as the RGB camera, and use HybrIK~\cite{li2021hybrik} as the 3D human pose estimator running at 30Hz. For the linear velocity estimation of the robot, we leverage the motion capture system (50Hz), and all the other proprioception is obtained from built-in sensors (200Hz) of Unitree H1 humanoid. Linear velocity state estimation could be replaced by onboard visual/LiDAR odometry methods, though we opt in to MoCap for this work due to its simplicity.

\begin{figure}[!htbp]
    \centering
    \includegraphics[width=0.5\textwidth]{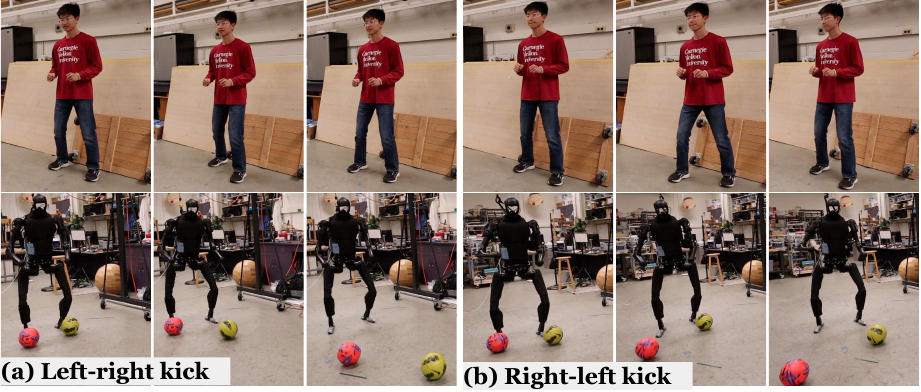}
    \caption{The humanoid robot is able to track the precise lower-body movements of the human teleoperator.}
    \label{fig:H2O_leftright}
    \vspace{-2mm}
\end{figure}

\begin{figure}[!htbp]
    \centering
    \includegraphics[width=0.5\textwidth]{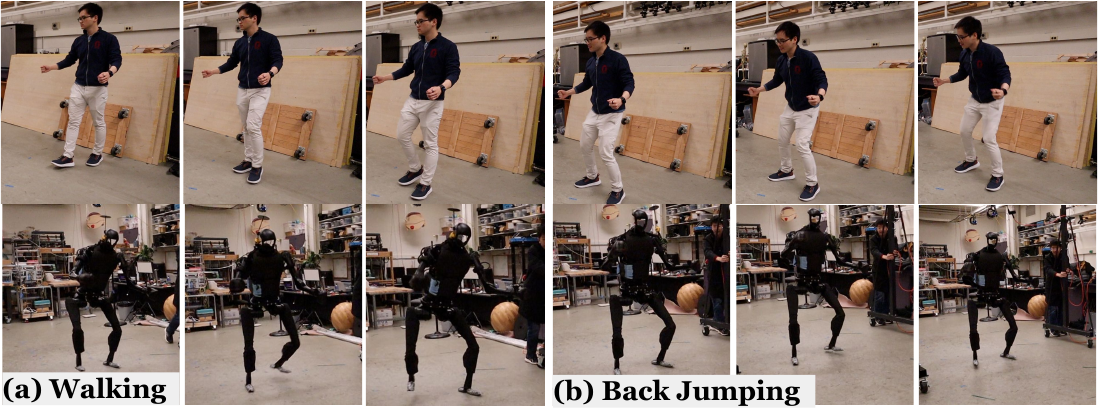}
    \caption{The humanoid robot is able to track walking motions of human-style pace and imitate continuous back jumping.}
    \label{fig:H2O_walk_backjumping}
    \vspace{-2mm}
\end{figure}

\paragraph{Real-world Teleoperation Results} For real-time teleoperation, the 3D pose estimation from the RGB camera is noisy and can suffer from perspective bias, but our \method policy shows a strong generalization ability to real-world estimated motion goals in real-time. The real-world teleoperation is shown in \Cref{fig:firstpage}, \Cref{fig:H2O_leftright} and \Cref{fig:H2O_walk_backjumping}, where \method enables precise real-time teleoperation of humanoids to do whole-body dynamic motions like ball kicking, walking, and back jumping. More demonstrations can be found on our \href{https://human2humanoid.com}{website}.

\begin{figure}[!htbp]
    \centering
    \includegraphics[width=0.5\textwidth]{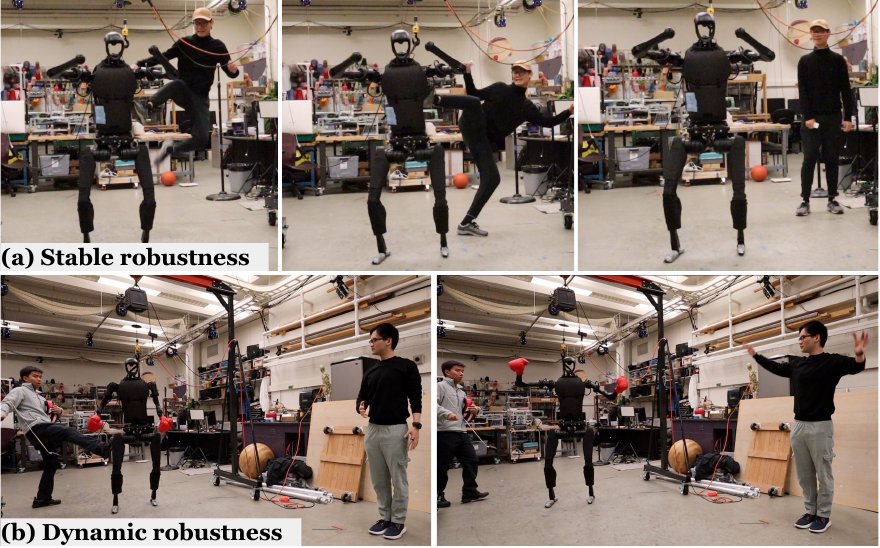}
    \caption{Robustness Tests of our \method system under powerful kicking. The policy is able to maintain balance for both stable and dynamic teleoperated motions.}
    \label{fig:H2O_robustness}
    \vspace{-4mm}
\end{figure}

\paragraph{Robustness}
Our \method system can keep balance under external force disturbances, as shown in \Cref{fig:H2O_robustness}. These tests demonstrate the robustness of our system.

\section{Discussions, Limitations, and Future Work}

\paragraph{Towards Universal Humanoid Teleoperation} 
Our ultimate goal is to enable the humanoid to follow as many human-demonstrated motions as possible. We emphasize three key factors that can be improved in the future. 
1) Closing the representation gap: as shown in \Cref{sec:simulaiton_experiment}, the state representation of the motion goals critically affects the scalability of RL training with more diverse motions, leading to a trade-off. While incorporating more expressive motion representations into the state space can accommodate finer-grained and more diverse motions, the expanded dimensionality will lead to a curse of sample efficiency in scalable RL. 
2) Closing the embodiment gap: as evident in \Cref{sec:simulaiton_experiment} and prior work~\cite{luo2024universal}, training on infeasible or damaged motions might largely harm performance. The feasibility of motions varies from robot to robot due to hardware constraints, and we lack systematic algorithms to identify feasible motions. We need more efforts to close this embodiment gap: on one end, more human-like humanoids would help; on the other, more teleoperation research is expected to improve the learnability of human motions. 
3) Closing the sim-to-real gap: to achieve a successful sim-to-real transfer, regularization (e.g., reward regularization) and domain randomization are needed. However, over-regularization and over-randomization will also hinder the policy from learning the motions. It remains unknown how to strike the best trade-off between motion imitation leaning and sim-to-real transfer into a universal humanoid control policy.

\paragraph{Towards Real-time Humanoid Teleoperation} In this work, we leverage RGB and 3D pose estimator to transform the motions of human teleoperators into humanoid robots. The latency and error from RGB cameras and pose estimation also lead to an inevitable trade-off between efficiency and precision in teleoperation. Also, in this work, the human teleoperator receives feedback from the humanoid only in the form of visual perception.
More research is needed on human-robot interaction to study this emerging multimodal interaction (e.g., force feedback~\cite{yang2022touch}, verbal and conversational feedback~\cite{chai2018language}), which could further enhance the capability of humanoid teleoperation.

\paragraph{Towards Whole-body Humanoid Teleoperation} 
One may wonder if lower-body tracking is necessary, as the major embodiment gap between humans and humanoids is the lower-body capability.
A large proportion of skillful motions of humans (e.g., sports, dancing) need diverse agile lower-body movements.
We emphasize the scenarios where legged robots hold an advantage over wheeled robots, in which lower-body tracking is necessary to follow human lower-body movements, including stepping stones, kicking, spread legs, etc. 
In the future, a teleoperated humanoid system that learns to switch between robust locomotion and skillful lower-body tracking would be a promising research direction.

\section{Conclusions}
In this study, we introduced Human to Humanoid (\method), a scalable learning-based framework that enables real-time whole-body humanoid robot teleoperation using just an RGB camera. Our approach, leveraging reinforcement learning and a novel ``sim-to-data" process, addresses the complex challenge of translating human motion into actions a humanoid robot can perform. Through comprehensive simulation and real-world tests, \method demonstrated its capability to perform a wide range of dynamic tasks with high fidelity and minimal hardware requirements.


\section*{ACKNOWLEDGMENT}
The authors express their gratitude to Jessica Hodgins for providing assistance in conducting hardware experiments. Special thanks are extended to Ziqiao Ma, Zhongyu Li, Yiyu Chen, Xuxin Cheng, and Unitree for their valuable help on graphics design and hardware debugging. Furthermore, we acknowledge the significance of CMU Wean Hall room 1334, formerly utilized as the recording location for the CMU MoCap dataset. In the present study, this dataset is used for real-world humanoid teleoperation within the same room.



\printbibliography

\end{document}